\pdfoutput=1

\documentclass[11pt]{article}

\usepackage[]{acl}

\usepackage{times}
\usepackage{latexsym}
\usepackage{bm}
\usepackage{hyperref}
\usepackage{color,colortbl}
\usepackage{framed}
\definecolor{shadecolor}{rgb}{0.92,0.92,0.92}
\usepackage{amsmath}

\usepackage{stfloats}
\usepackage{enumitem}
\usepackage{booktabs}
\usepackage{nccmath}
\usepackage{multirow}
\usepackage{subfigure}
\usepackage{latexsym}
\usepackage{todonotes}
\usepackage{makecell}
\usepackage{dashrule}
\usepackage{amssymb}
\usepackage{bbding}

\usepackage{graphicx}

\definecolor{LightGray}{gray}{0.9}
\usepackage{soul}

\usepackage{arydshln}

\usepackage{amssymb}
\usepackage{pifont}
\usepackage[T1]{fontenc}

\usepackage[utf8]{inputenc}

\usepackage{microtype}

\usepackage{soul}

\title{Is ChatGPT a Good Multi-dimensional Evaluator for Text Style Transfer?}

\title{Multidimensional Evaluation for Text Style Transfer Using ChatGPT}

\author{
Huiyuan Lai, Antonio Toral, Malvina Nissim\\
CLCG, University of Groningen / The Netherlands\\
\texttt{\{h.lai, a.toral.ruiz, m.nissim\}@rug.nl}
}

\begin{document}
\maketitle
\begin{abstract}

We investigate the potential of ChatGPT as a multidimensional evaluator for the task of \emph{Text Style Transfer}, alongside, and in comparison to, existing automatic metrics as well as human judgements. We focus on a zero-shot setting, i.e.\ prompting ChatGPT with specific task instructions, and test its performance on three commonly-used dimensions of text style transfer evaluation: style strength, content preservation, and fluency. We perform a comprehensive correlation analysis for two transfer directions (and overall) at different levels.
Compared to existing automatic metrics, ChatGPT achieves competitive correlations with human judgments. These preliminary results are expected to provide a first glimpse into the role of large language models in the multidimensional evaluation of stylized text generation. 

\end{abstract}

\section{Introduction}

Large Languages Models (LLMs) have demonstrated 
a
remarkable ability to solve Natural Language Processing (NLP) tasks in a zero-shot manner, using task-specific instructions instead of requiring task-specific training~\citep{brown2020language, chowdhery2022palm}. 
Particularly, the LLM ChatGPT\footnote{\url{https://openai.com/blog/chatgpt}} developed by OpenAI has garnered a great deal of attention from both the academic and industry communities, as well as general users. This model achieves impressive results across a wide range of NLP tasks, such as text summarization~\citep{yang2023exploring, qin2023chatgpt}, question answering~\citep{omar2023chatgpt, qin2023chatgpt}, mathematical reasoning~\cite{frieder2023mathematical}, and machine translation~\citep{jiao2023chatgpt}. These achievements are attributed to the model's strong ability to complete the target task in a human-like conversational manner.

Recent studies have investigated if ChatGPT can not only serve as a text generator, but also as an \emph{evaluator} for other systems performing NLP tasks, such as machine translation~\citep{kocmi2023large} and several general text generation tasks~\citep{wang2023chatgpt}. These studies compare ChatGPT to existing automatic metrics (e.g.\ ROUGE,~\citet{lin-2004-rouge}), assessing correlation with human judgments. While these works show that ChatGPT can provide more reliable Natural Language Generation (NLG) evaluations, they mostly focus on content (meaning) and fluency aspects.

In our study, we extend the evaluation of ChatGPT to a popular controllable text generation task, namely \textit{Text Style Transfer} (TST), not only examining content and fluency, but also investigating  the evaluation of the specifically controlled text property, namely style. We also compare to a wider range of state-of-the art automatic metrics than done before.

TST is the task of transforming a text of one style into another while preserving its style-independent content. When evaluating the transformed sentence, three dimensions are commonly assessed: (1) the content of the transformed sentence is the same as the source sentence (\textit{content preservation}); (2) the transformed sentence fits the target style (\textit{style strength}); and (3) the transformed sentence is readable and fluent for the target style (\textit{fluency}). 

Human judgements can often provide reliable assessments of TST systems, but they are costly and not easy to employ during iterative system development. Hence, researchers have dedicated many efforts to developing various automatic metrics with the aim of reducing the need to rely on human judgements. This strategy implies developing separate metrics for each dimension (for example, a style classifier to assess style strength, or training neural metrics such as COMET \cite{rei-etal-2020-comet} to evaluate content preservation), with the additional burden of having to (re)train such metrics on domain-specific data. For example, to measure perplexity when assessing fluency, different models have to be specialised so that the target style is well represented; similarly, a classifier to evaluate style strength has to be trained using style-specific data and is not directly portable to other TST tasks.

Having a single model that is able to provide, without training, evaluation for all three dimensions would then be a substantial step forward, especially as support during development. Thus, in this contribution we address this research question: ``\textit{Can ChatGPT be reliably used as a single, multidimensional TST evaluator in a zero-shot manner?}''

Specifically, we focus on \textit{formality transfer}, a common TST task \citep{rao-tetreault-2018-dear} where the aim is to transform a formal source sentence into an informal counterpart, or vice versa (e.g.\ informal: \textit{my daughter is super into it and i think it's just so darn cute!!} $\leftrightarrow$ formal: \textit{My daughter is quite passionate about it, and I find it adorable.}). We design various task instructions to prompt ChatGPT to score the outputs of multiple existing TST systems across the three different dimensions mentioned above and run correlations with human judgements. Our main contributions are as follows:

\begin{itemize}[leftmargin=*]
\itemsep 0in

\item We present the first investigation of ChatGPT as a TST evaluator through the analysis of correlations with human judgements.

\item We experiment with 16 automatic evaluation metrics and suggest that ChatGPT has the potential to perform multidimensional TST evaluation.

\item We experiment with different prompt templates, showing that a simple template with separate requests for the different evaluation dimensions achieves the best performance. 

\end{itemize}

\noindent We release all the code and data, including prompt templates used for the experiments, as well as the corresponding scoring results.\footnote{\url{https://github.com/laihuiyuan/eval-formality-transfer}.}

\section{Background}

\subsection{ChatGPT}

ChatGPT is a chatbot released by OpenAI, which is built upon LLM GPT-$\ast$ family model~\citep{brown2020language}. Due to its impressive performance, ChatGPT has attracted over 100 million subscribers 2.5 months after its launch and has been described as the fastest-growing web platform ever~\citep{haque2022i, leiter2023chatgpt}. With the algorithm of reinforcement learning from human feedback (RLHF), ChatGPT is trained to interact with users in a conversational way. By providing prompts as guidance to trigger specific tasks' capabilities, users can get ChaGPT to generate task-related responses without any task-specific training data. Relatedly, and expectedly, recent work has shown that the type and quality of the generated outputs are substantially affected by the specific prompt(s) used~\citep{jiao2023chatgpt, kocmi2023large}.

\subsection{TST Evaluation}

TST systems are assessed using various metrics for the three evaluation dimensions: style, content, and fluency. For instance, BLEU~\citep{papineni-etal-2002-bleu} is the most popular content-related metric in previous work~\citep{briakou-etal-2021-evaluating, lai-etal-2022-human}. In recent years, many studies have examined a range of automatic methods in the context of TST~\citep{rao-tetreault-2018-dear, yamshchikov2020style, briakou-etal-2021-evaluating, lai-etal-2022-human}. A valid metric should express the quality of the output, and correlation with human judgements is taken as a reliable indicator of the metric's validity. Here, we perform a rich correlation analysis to assess whether ChatGPT can be used as a multidimensional TST evaluator across the three aforementioned evaluation dimensions.

\subsection{TST Systems}
Like in most NLG tasks, models trained with parallel data can provide a stronger foundation for TST~\citep{rao-tetreault-2018-dear, lai-etal-2021-thank}. This has prompted a series of TST approaches, including data augmentation~\citep{zhang-etal-2020-parallel}, multitask learning~\citep{niu-etal-2018-multi}, reinforced learning~\citep{Abhilasha-2020}, and augmenting pre-trained models with rewards~\citep{lai-etal-2021-thank}. Since parallel data is often unavailable, though, a substantial amount of work has gone into exploring unsupervised approaches: disentangling style and content~\citep{shen2017style, john-etal-2019-disentangled, yi-2020-text}, style-word-editing~\citep{li-etal-2018-delete, wu-etal-2019-hierarchical-reinforced, lee-2020-stable}, utilizing generic resources~\citep{chawla-yang-2020-semi, lai-etal-2021-generic}, back-translation~\citep{lample2018multipleattribute, luo-2019-adual}, language and task adaptation~\citep{lai-etal-2022-multilingual}. The TST systems we consider in this paper exemplify various of such approaches \citep{lai-etal-2022-human}.

\section{ChatGPT for TST Evaluation}

We create a ChatGPT evaluation request for each source sentence and its corresponding system outputs, conducting experiments in a zero-shot manner with task instructions.
Concretely, we give instructions to ChatGPT in the same way that we would do to a human evaluator, providing it with task-specific instructions, which mainly include (1) specific transformation directions and (2) the dimensions that should be evaluated. As an example, we show a prompt template for evaluating content preservation in the informal-to-formal transformation, as shown in Figure~\ref{fig:template}.

\begin{figure}[!t]
\centering
\begin{minipage}{0.9\linewidth}
\begin{shaded}
\noindent For the task of \textcolor[RGB]{101,42,150}{\texttt{informal-to-formal style transfer}}, score the \textcolor[RGB]{101,42,150}{\texttt{content preservation}} of each output with respect to the source sentence on a scale of 0 to 100, with 100 being complete preservation of the original meaning, and 0 being completely incorrect or altered meaning. The evaluation format is as follows:\\

\noindent Output 1 \\
\noindent Score:\\
\noindent Explanation:\\
$---------------------$\\
\noindent Source text: []\\
\noindent Output 1: []\\
\noindent ...\\
\noindent Output n: []
\end{shaded}
\end{minipage}
\caption{Prompt template for the evaluation on content preservation.}
\label{fig:template} 
\end{figure}

After prompting ChatGPT with the task instruction, we provide it with the source sentence and its corresponding outputs produced by various systems we consider. In this way, ChatGPT is guided to generate the assessment of the content aspect (e.g.\ ``Score: 100'') for each output, also providing a corresponding explanation (e.g.\ ``Explanation: The output is almost identical to the source sentence and preserves the original meaning completely.''). We experiment with three evaluation dimensions (content preservation, style strength, and fluency) in two transfer directions (informal-to-formal: I$\rightarrow$F, and formal-to-informal: F$\rightarrow$I) and overall.\footnote{Complete prompt templates are in Appendix~\ref{app:prompt}.}

\section{Experiments}

\subsection{Data and Baselines}
 
 We perform the experiments using the formality transfer data released by~\citet{lai-etal-2022-human}. This dataset contains 80 source sentences (40 for each transfer direction), as well as their corresponding 9 outputs (8 systems + 1 human reference). Each output has 6 human judgments and 16 different automatic evaluation results covering three evaluation dimensions: content preservation, style strength, and fluency. Details are as follows.

\paragraph{Content Preservation}  There are 8 different metrics: BLEU~\citep{papineni-etal-2002-bleu}, chrF~\citep{popovic-2015-chrf},  ROUGE~\citep{lin-2004-rouge}, WMD~\citep{pmlr-v37-kusnerb15}, METEOR~\citep{banerjee-lavie-2005-meteor}, BERTScore~\citep{bert-score}, BLEURT~\citep{sellam-etal-2020-bleurt} and COMET~\citep{rei-etal-2020-comet}.

\paragraph{Style Strength} We train three models based on BERT~\citep{devlin-etal-2019-bert}: two style classifiers (C-GYAFC and C-PT16), each trained with a different data set, and a style regressor (R-PT16).

\paragraph{Fluency} The language model GPT-2~\citep{radford-2019} is further pre-trained on style labelled texts; the resulting (in)formal models are then used to calculate the perplexity of the generated sentence in the corresponding target style.

\subsection{Correlation Methods}


Given a set of source texts (sentences in our case) $\{t_1, t_2, \dots, t_N\}$ and a set of style transfer systems $\{s_1, s_2, \dots, s_M\}$, 
the generated text of the $j$-th model for the $i$-th source text is denoted as $g_{i,j}$. To measure how well automatic metrics correlate with human judgements, we calculate their correlation scores at three levels.

\paragraph{System-level} Following previous work~\citep{kocmi2023large}, we adopt pairwise accuracy (Acc, ~\citet{kocmi-etal-2021-ship}) as an overall measure to assess how well automatic metrics correlate with humans at \textit{system level}. Specifically, 
given all pairs of systems, the method checks how many of them are ranked in the same way (e.g. system 1 is better than system 2) by both the automatic metric and the human ranking and divides that by the total number of pairs. We average sentence-level scores to obtain system-level scores for each transfer direction.
Besides the correlations for each direction, we also 
calculate the overall correlation, by averaging the correlations for the two directions.

\begin{equation}
\small
\begin{split}
     \textrm{Acc} = \frac{|\textrm{sign}(\textrm{metric} \Delta) = \textrm{sign}(\textrm{human} \Delta)|}{|\textrm{all system pairs}|}
\end{split}
\end{equation}

\noindent Where $\Delta$ is the difference in
metric (or human) scores between two systems.

\paragraph{Sample-level} With this method a correlation score is calculated for each individual source sentence separately based on outputs of multiple systems, then averaged across all scores.

\begin{equation}
\small
\begin{split}
    \textrm{C}_{sample} = \frac{1}{N} \sum_{i=1}^{N} (\rho ([f_{a}(g_{i,1}), ..., f_{a}(g_{i,M})],\\
    [f_{h}(g_{i,1}), ..., f_{h}(g_{i,M})]))
\end{split}
\end{equation}

\noindent  Where $f_a$ and $f_h$ are the scores of automatic metric and human judgment, respectively. $\rho$ denotes the correlation metric and we adopt Kendall’s Tau~\citep{maurice1938}, which we also use for the dataset-level correlation below.

\paragraph{Dataset-level} It defines the correlation score calculated on system outputs of all $n$ samples.

\begin{equation}
\small
\begin{split}
    \textrm{C}_{dataset} = \rho ([f_{a}(g_{1,1}), \dots, f_{a}(g_{n,M})], \\
    [f_{h}(g_{1,1}), \dots, f_{h}(g_{n,M})])
\end{split}
\end{equation}

\noindent This method takes a holistic view of all output samples. By exploiting the comparison of the outputs from different models for all source sentences, it solely concentrates on the quality of the sentences.

\begin{table*}[!t]
\footnotesize
\centering
\begin{tabular}{lccccccccc}
\toprule
& \multicolumn{3}{c}{System-level} & \multicolumn{3}{c}{Sample-level} & \multicolumn{3}{c}{Dataset-level}\\
  \cmidrule(lr){2-4}\cmidrule(lr){5-7}\cmidrule(lr){8-10}
              & I$\rightarrow$F & F$\rightarrow$I & Overall & I$\rightarrow$F & F$\rightarrow$I & Overall & I$\rightarrow$F & F$\rightarrow$I & Overall\\
 \midrule
 IAA          & 97.2 & 97.2 & 97.2 & 59.9 & 60.0 & 59.9 & 48.0 & 50.6 & 49.5\\
 \hdashline
 BLEU         & 47.2 & 69.4 & 58.3 & 24.4 & 39.2 & 31.8 & 17.0 & 32.8 & 24.8\\  
 ROUGE-1      & 38.9 & 75.0 & 57.0 & 15.4 & 35.0 & 25.2 & 14.4 & 35.7 & 24.2\\
 ROUGE-2      & 41.7 & 72.2 & 57.0 & 20.6 & 37.2 & 28.9 & 16.3 & 33.3 & 24.8\\
 ROUGE-L      & 38.9 & 72.2 & 55.6 & 16.1 & 35.9 & 26.0 & 14.4 & 34.5 & 24.3\\
 ChrF         & 75.0 & 80.6 & 77.8 & 37.0 & 42.1 & 39.5 & 29.4 & 41.4 & 34.6\\
 WMD          & 58.3 & 77.8 & 68.1 & 25.0 & 42.3 & 33.7 & 20.2 & 39.2 & 28.9\\
 METEOR       & 61.1 & 77.8 & 69.5 & 37.3 & 40.8 & 39.0 & 23.1 & 31.8 & 28.1\\
 BERTScore    & 80.6 & 91.7 & 86.2 & 47.4 & 56.1 & 51.7 & 36.8 & 54.4 & 45.8\\
 BLEURT       & \textbf{94.4} & \textbf{94.4} & \textbf{94.4} & \textbf{68.7} & 62.3 & \textbf{65.5} & 48.0 & 57.5 & 52.9\\
 COMET-20     & 91.7 & \textbf{94.4} & 93.1 & 56.7 & 61.6 & 59.1 & 46.6 & 58.5 & 52.1\\
 COMET-22     & 91.7 & 88.9 & 90.3 & 57.8 & 60.2 & 59.0 & 44.4 & 56.7 & 50.0\\
 ChatGPT      & 80.6 & \textbf{94.4} & 87.5  & 46.8 & \textbf{64.9} & 55.9 & \textbf{48.3} & \textbf{60.6} & \textbf{54.3}\\
\bottomrule
\end{tabular}
\caption{\label{tab:corr-content-ref}
 Correlation results in terms of content preservation. Highest correlation per column shown in bold.}
 \end{table*}

\begin{table*}[!t]
\footnotesize
\centering
\begin{tabular}{lccccccccc}
\toprule
& \multicolumn{3}{c}{System-level} & \multicolumn{3}{c}{Sample-level} & \multicolumn{3}{c}{Dataset-level}\\
  \cmidrule(lr){2-4}\cmidrule(lr){5-7}\cmidrule(lr){8-10}
              & I$\rightarrow$F & F$\rightarrow$I & Overall & I$\rightarrow$F & F$\rightarrow$I & Overall & I$\rightarrow$F & F$\rightarrow$I & Overall\\
 \midrule
 IAA       & 88.9 & 72.2 & 80.6 & 58.5 & 15.4 & 36.9 & 44.0 &  1.9 & 28.1\\
 \hdashline 
 R-PT16    & \textbf{91.7} & 47.2 & 69.5 & 68.0 & -7.9 & 30.0 & 48.2 & -11.0 & 13.0\\
 C-PT16    & 86.1 & 58.3 & 72.2 & 33.9 &  3.1 & 18.5 & 29.1 &  10.1 & 20.9\\
 C-GYAFC   & \textbf{91.7} & 41.7 & 66.7 & \textbf{76.3} &  6.8 & 41.5 & \textbf{59.0} &  4.4  & \textbf{40.3}\\
 ChatGPT   & 86.1 & \textbf{66.7} & \textbf{76.4} & 63.1 & \textbf{30.8} & \textbf{41.9} & 47.2 &  \textbf{16.6} & 31.2\\
\bottomrule
\end{tabular}
\caption{\label{tab:corr-style-ref}
 Correlation results in terms of style strength. Highest correlation per column shown in bold.}
 \end{table*}

\section{Results and Analysis}

\subsection{Main results}

We include here our correlation and inter-annotator agreement (IAA) results for the 8 systems, one reference, and human judgements that are provided by \citet{lai-etal-2022-human}. Results with the 8 systems (no reference included) can be found in Appendix~\ref{app:corr-8}.

\paragraph{Content Preservation} We conduct experiments on this dimension using source sentences, aiming to evaluate content preservation when transforming source into target even in the absence of a human-written reference. Specifically, we calculate the correlation of automatic metrics (computed against source sentences) with human judgments; results are shown in Table~\ref{tab:corr-content-ref}.

We see that for the three levels, the inter-annotator agreement scores are very close in both transfer directions and also overall. For system-level, ChatGPT achieves the best correlation together with BLEURT and COMET-20 in the F$\rightarrow$I direction, while BLEURT achieves the highest scores in the other direction and overall.
We can also observe that ChatGPT performs the best in the 
F$\rightarrow$I direction at the sample level, but is behind the best-performing metric BLEURT in the I$\rightarrow$F direction and overall. This may be because ChatGPT cannot ensure evaluation consistency across all transferred sentences, i.e. two systems with the same output can get different scores. It is interesting to see that ChatGPT yields the best dataset-level scores in both transfer directions and overall.

\paragraph{Style Strength} Table~\ref{tab:corr-style-ref} presents the inter-annotator agreement and correlation of automatic methods in terms of style strength with human judgements in each direction and overall. We see that scores for the I$\rightarrow$F transfer are always higher than those in the opposite direction. In line with this, the results of inter-annotator agreement (IAA) show that humans are more subjective in their judgments about informality. Regarding automatic methods, we can observe that: (i) they correlate well with human judgements in the I$\rightarrow$F direction, except C-PT16 which has lower scores at the sample level and dataset level; (ii) ChatGPT performs the best in the F$\rightarrow$I direction while some other methods like R-PT16 yield negative correlation scores at the sample level and dataset level; (iii) C-PT16 has the highest overall score at the dataset level, while ChatGPT performs the best at system and sample level.

 \begin{table*}[!t]
\footnotesize
\centering
\begin{tabular}{lccccccccc}
\toprule
& \multicolumn{3}{c}{System-level} & \multicolumn{3}{c}{Sample-level} & \multicolumn{3}{c}{Dataset-level}\\
  \cmidrule(lr){2-4}\cmidrule(lr){5-7}\cmidrule(lr){8-10}
              & I$\rightarrow$F & F$\rightarrow$I & Overall & I$\rightarrow$F & F$\rightarrow$I & Overall & I$\rightarrow$F & F$\rightarrow$I & Overall\\
 \midrule
 IAA      & 97.2 & 86.2 & 91.7 & 50.5 & 40.2 &  5.4 & 46.8 & 30.0 & 39.0\\
 \hdashline
 GPT-2    & \textbf{83.3} & 80.6 & 82.0 & 44.6 & 29.8 & 37.7 & 37.7 & 26.7 & 32.0\\
 ChatGPT  & \textbf{83.3} & \textbf{83.3} & \textbf{83.3} & \textbf{64.8} & \textbf{37.4} & \textbf{51.1} & \textbf{52.3} & \textbf{35.2} & \textbf{42.5}\\
\bottomrule
\end{tabular}
\caption{\label{tab:corr-fluency-ref}
 Correlation results in terms of fluency. Highest correlation per column shown in bold.}
 \end{table*}

\paragraph{Fluency} Table~\ref{tab:corr-fluency-ref} presents the correlation results in terms of fluency. Similar to style strength, the score of I$\rightarrow$F, including IAA, is always higher than the score in the opposite direction. This suggests that human perceptions of fluency are more diverse in informal sentences than in formal ones. Unsurprisingly, we see that ChatGPT correlates better with human judgments than GPT-2 on all three levels 
despite GPT-2 being fine-tuned with domain data. At system level, there is also no difference between the two directions.

 \begin{table}[!t]
\footnotesize
\centering
\begin{tabular}{llccc}
\toprule
 Level & Dimension & Content & Style & Fluency\\
 \midrule
 \multirow{2}{*}{System} & Single    & \textbf{87.5}  & 76.4 & 83.3\\
                         & Multiple  & \textbf{87.5} & \textbf{77.8} & \textbf{91.7}\\
 \hline
 \multirow{2}{*}{Sample} & Single    & \textbf{55.9} & \textbf{41.9} & \textbf{51.1}\\
                         & Multiple  & 48.0 & 41.4 & 50.3\\
 \hline
 \multirow{2}{*}{Dataset}& Single    & \textbf{54.3} & \textbf{31.2} & \textbf{42.5}\\
                         & Multiple  & 48.7 & 26.3 & 37.6\\
\bottomrule
\end{tabular}
\caption{\label{tab:com-results}
 Overall correlation results of ChatGPT with different prompts. Highest correlation per level and column shown in bold.}
 \end{table}

\subsection{Impact of Prompts}

We experiment with a slightly different prompt configuration: for a given source sentence and corresponding outputs, ChatGPT is asked to provide assessments on the three evaluation dimensions of the outputs \textit{at once}, instead of one at a time. Here, we reuse the human annotation guidelines provided by~\citet{lai-etal-2022-human} to prompt ChatGPT.\footnote{The template example is in Appendix~\ref{app:prompt}.} The guidelines contains very detailed information on the definition of style transfer and its evaluation on different dimensions.

As shown in Table~\ref{tab:com-results}, we see that prompting ChatGPT with instructions of a single evaluation dimension performs better in most settings, except system-level style and fluency.
Based on word count, the prompt for a single evaluation dimension is the least constrained template (content: 54 and style/fluency: 52 vs multiple dimensions: 188). 
Our finding is consistent with~\citet{kocmi2023large} that the least constrained template often achieves the best performance.

\section{Conclusion}

We investigated the potential of ChatGPT as a
evaluator for text style transfer from multiple dimensions: content preservation, style strength, and fluency. By prompting ChatGPT with specific evaluation instructions, we found that it achieves competitive correlations with human judgements compared to existing automatic metrics. This is particularly the case for the most difficult setting evaluated (dataset-level), on which ChatGPT outperform the other metrics in almost all cases.
This suggests that ChatGPT has the potential to be a multidimensional evaluator for text style transfer.

\begin{table*}[!t]
\footnotesize
\centering
\begin{tabular}{lp{12cm}}
\toprule
Type & Example\\
\hline
\multirow{5}{*}{Refuse to evaluate} & 
   Source: Not all gentlemen break up with their woman.\\
 & Output 1: Not all guys break up with their girl.\\
 \\
 & ChatGPT: I'm sorry, but I cannot provide informal strength scores for Output 1 as it includes an offensive term. It's important to be mindful of using appropriate and respectful language.\\
\hline
\multirow{11}{*}{\makecell[l]{Inconsistent assessment\\ (content preservation)}} & 
 Source: that way you can still go by your last name!\\\\
 
 & Output 2: That way you can still go by your last name.\\
 & Score: 90\\
 & Explanation: The output sentence is similar to the source sentence but could be improved by adding a comma after "way" to improve readability.\\\\
 
 & Output 5: That way you can still go by your last name.\\
 & Score: 100\\
 & Explanation: Explanation: The output sentence preserves the original meaning and conveys the same message in a more formal style. The sentence structure and wording are different, but the content is accurately conveyed.\\
\bottomrule
\end{tabular}
\caption{\label{tab:example}
 Examples of assessments produced by ChatGPT.}
 \end{table*}


\section{Limitations}

Leveraging available data from previous work, we focus on formality style transfer by performing correlation analysis with human judgments to provide the understanding of ChatGPT in automatic evaluation on this task. It is important to note that the findings here may not generalise to other tasks usually considered similar, such as sentiment swap. We only experiment with two prompt templates in a zero-shot setting; future work such explore few-shot prompting, too. Additionally, as shown in Table~\ref{tab:example}, we observed some intrinsic limitations in using ChatGPT such that  (i) it sometimes refuses to evaluate cases that contain offensive text; (ii) it cannot ensure evaluation consistency across sentences in our experiments as it may return different assessments for identical outputs. A direction to explore is to design a specific prompt to explicitly guide it in keeping the evaluation (more) consistent. 
Finally, we conceive this as a first step to test how well conversational agents based on generative language models can serve the purpose of evaluating automatically generated output in the context of TST. Future work will need to extend testing to other ChatGPT-like models, possibly more open, with different sizes and languages.

\bibliography{anthology,custom}
\bibliographystyle{acl_natbib}

\clearpage

\appendix
\onecolumn
\section{Appendix \\~ \\ }
\label{sec:appendix}

\subsection{Prompt Templates}
\label{app:prompt}

\begin{figure}[!ht]
\centering
\begin{minipage}{0.95\linewidth}
\begin{shaded}
\noindent For the task of \textcolor[RGB]{101,42,150}{\texttt{informal-to-formal style transfer}}, score the \textcolor[RGB]{101,42,150}{\texttt{formal strength}} of each output on a scale of 0 to 100, with 100 being a perfect representation of formal writing style and 0 being informal or not adhering to formal writing style. The evaluation format is as follows:\\

\noindent Output 1\\
\noindent Score:\\
\noindent Explanation:

\end{shaded}
\end{minipage}

\caption{Prompt template for the evaluation on style strength.}
\label{fig:template-style} 
\end{figure}

\begin{figure}[!ht]
\centering

\begin{minipage}{0.95\linewidth}
\begin{shaded}
\noindent For the task of \textcolor[RGB]{101,42,150}{\texttt{informal-to-formal style transfer}}, score the \textcolor[RGB]{101,42,150}{\texttt{fluency}} of each output with respect to the source sentence on a scale of 0-100 in the following format, with 100 being extremely fluent and 0 being extremely difficult to read or understand. The evaluation format is as follows:\\

\noindent Output 1\\
\noindent Score:\\
\noindent Explanation:
\end{shaded}
\end{minipage}

\caption{Prompt template for the evaluation on fluency.}
\label{fig:template-fluency} 
\end{figure}

\begin{figure}[!ht]
\centering
\begin{minipage}{0.95\linewidth}
\begin{shaded}
\noindent We are conducting research on the use of English, especially on formality. We would like you to evaluate the sentences given, specific instructions for evaluation are provided in the following:\\

\noindent This task consists of judging sentence changes. To this aim, you will be shown different changes in a given sentence. The changes are related to style: from informal to formal or from formal to informal. We call these changes transformations.\\

\noindent You are asked to judge to what extent the transformations are successful by assessing the content (The content of the transformed sentence is the same as the source sentence.), the style (The transformed sentence fits the target style.), and the fluency (Considering the target style, the transformed sentence could have been written by a native speaker.) of the new sentence. You will see an original sentence and various possible transformations. For each transformation, you will have to use a continuous scale score from 0 to 100 to indicate how much you agree with each statement. The higher score you give, the higher your level of agreement. The evaluation format is as follows:\\

\noindent Output 1\\
\noindent Content: \\
\noindent Style: \\
\noindent Fluency: \\
\noindent Explanation: 
\end{shaded}
\end{minipage}

\caption{Prompt template for multidimensional evaluation.}
\label{fig:template-multi} 
\end{figure}

\clearpage

\subsection{Correlation Results on 8 Systems}
\label{app:corr-8}

\begin{table*}[!ht]
\footnotesize
\centering
\begin{tabular}{lccccccccc}
\toprule
& \multicolumn{3}{c}{System-level} & \multicolumn{3}{c}{Sample-level} & \multicolumn{3}{c}{Dataset-level}\\
  \cmidrule(lr){2-4}\cmidrule(lr){5-7}\cmidrule(lr){8-10}
              & I$\rightarrow$F & F$\rightarrow$I & Overall & I$\rightarrow$F & F$\rightarrow$I & Overall & I$\rightarrow$F & F$\rightarrow$I & Overall\\
 \midrule
 IAA          & 96.4 & 96.4 & 96.4 & 57.0 & 60.0 & 58.5 & 50.5 & 51.4 & 51.2\\
 \hdashline
 BLEU         & 42.9 & 71.4 & 57.2 & 22.9 & 42.6 & 32.8 & 17.6 & 33.1 & 24.8\\  
 ROUGE-1      & 32.1 & 78.6 & 55.4 & 13.1 & 35.6 & 24.4 & 14.5 & 35.5 & 23.7\\
 ROUGE-2      & 35.7 & 75.0 & 55.4 & 19.5 & 39.2 & 29.3 & 16.8 & 33.6 & 24.6\\
 ROUGE-L      & 32.1 & 75.0 & 53.6 & 14.2 & 37.2 & 25.7 & 14.9 & 35.6 & 24.1\\
 ChrF         & 78.6 & 85.7 & 82.2 & 38.7 & 44.0 & 41.4 & 31.2 & 42.8 & 35.7\\
 WMD          & 57.1 & 82.1 & 69.6 & 23.4 & 44.1 & 33.8 & 21.4 & 39.0 & 29.1\\
 METEOR       & 60.7 & 82.1 & 71.4 & 38.1 & 44.8 & 41.4 & 24.9 & 32.2 & 28.9\\
 BERTScore    & 85.7 & 100  & 92.9 & 49.3 & 59.3 & 54.3 & 38.8 & 55.8 & 47.4\\
 BLEURT       & 92.9 & 100  & 96.5 & 67.9 & 64.0 & 65.9 & 50.3 & 58.1 & 54.1\\
 COMET-20     & 89.3 & 100  & 94.7 & 57.6 & 63.8 & 60.7 & 48.7 & 59.1 & 53.3\\
 COMET-22     & 89.3 & 92.9 & 91.1 & 58.3 & 61.1 & 59.7 & 46.4 & 56.3 & 50.7\\
 ChatGPT      & 75.0 & 96.4 & 85.7  & 46.1 & 64.6 & 55.3 & 49.7 & 60.5 & 54.6\\
\bottomrule
\end{tabular}
\caption{\label{tab:corr-content}
 Correlation results in terms of content preservation at three levels. }
 \end{table*}

\begin{table*}[!ht]
\footnotesize
\centering
\begin{tabular}{lccccccccc}
\toprule
& \multicolumn{3}{c}{System-level} & \multicolumn{3}{c}{Sample-level} & \multicolumn{3}{c}{Dataset-level}\\
  \cmidrule(lr){2-4}\cmidrule(lr){5-7}\cmidrule(lr){8-10}
              & I$\rightarrow$F & F$\rightarrow$I & Overall & I$\rightarrow$F & F$\rightarrow$I & Overall & I$\rightarrow$F & F$\rightarrow$I & Overall\\
 \midrule
 IAA       & 92.9 & 71.4 & 82.2 & 61.7 & 15.5 & 38.6 & 44.3 & 1.8 & 29.1\\
 \hdashline 
 R-PT16    & 92.9 & 57.1 & 75.0 & 71.8 & -6.0 & 32.9 & 47.5 & -8.4 & 14.1\\
 C-PT16    & 85.7 & 50.0 & 67.9 & 73.6 &  3.8 & 38.7 & 42.7 & 11.3 & 26.6\\
 C-GYAFC   & 89.3 & 46.4 & 67.9 & 75.2 &  7.6 & 41.4 & 53.9 &  8.4 & 36.2\\
 ChatGPT   & 89.3 & 71.4 & 80.4 & 68.9 & 15.9 & 42.4 & 49.0 & 14.2 & 31.2\\
\bottomrule
\end{tabular}
\caption{\label{tab:corr-style}
 Correlation results in terms of style strength at three levels. }
 \end{table*}

\begin{table*}[!ht]
\footnotesize
\centering
\begin{tabular}{lccccccccc}
\toprule
& \multicolumn{3}{c}{System-level} & \multicolumn{3}{c}{Sample-level} & \multicolumn{3}{c}{Dataset-level}\\
  \cmidrule(lr){2-4}\cmidrule(lr){5-7}\cmidrule(lr){8-10}
              & I$\rightarrow$F & F$\rightarrow$I & Overall & I$\rightarrow$F & F$\rightarrow$I & Overall & I$\rightarrow$F & F$\rightarrow$I & Overall\\
 \midrule
 IAA      & 96.4 & 85.7 & 91.1 & 55.6 & 42.9 & 49.3 & 49.4 & 32.3 & 41.7\\
 \hdashline
 GPT-2    & 89.3 & 78.6 & 84.0 & 50.9 & 35.3 & 43.1 & 41.6 & 28.7 & 34.9\\
 ChatGPT  & 85.7 & 82.1 & 83.9 & 67.0 & 41.9 & 55.0 & 54.7 & 38.7 & 45.6\\
\bottomrule
\end{tabular}
\caption{\label{tab:corr-fluency}
 Correlation results in terms of fluency at three levels. }
 \end{table*}

\end{document}